\documentclass{article}

\usepackage[final]{neurips_2026}
\makeatletter
\providecommand{\@trackname}{}
\makeatother

\usepackage[utf8]{inputenc}
\usepackage[T1]{fontenc}
\usepackage{hyperref}
\usepackage{url}
\usepackage{booktabs}
\usepackage{multirow}
\usepackage{graphicx}
\usepackage{amsfonts}
\usepackage{bbm}
\usepackage{nicefrac}
\usepackage{microtype}
\usepackage{comment}
\usepackage[table]{xcolor}
\usepackage{amsmath}

\newcommand{\cb}{{\boldsymbol{c}}}

\newcommand{\xb}{{\boldsymbol{x}}}
\newcommand{\yb}{{\boldsymbol{y}}}

\newcommand{\vb}{{\boldsymbol{v}}}

\newcommand{\epsilonb}{{\boldsymbol{\epsilon}}}

\newcommand{\Ac}{{\mathcal{A}}}

\newcommand{\Nc}{{\mathcal{N}}}
\newcommand{\Mc}{{\mathcal{M}}}

\newcommand{\Xb}{{\boldsymbol{X}}}

\usepackage{wrapfig}      
\usepackage{colortbl}     
\usepackage{xcolor}   
\usepackage{enumitem}
\setlist[itemize]{leftmargin=1.2em, itemsep=0pt, topsep=2pt}
\usepackage{caption}
\usepackage{titlesec}
\titlespacing*{\section}{0pt}{8pt plus 2pt minus 2pt}{4pt}
\titlespacing*{\subsection}{0pt}{6pt plus 2pt minus 2pt}{3pt}

\newcommand{\best}[1]{\cellcolor{blue!30}\textbf{#1}}
\newcommand{\second}[1]{\cellcolor{blue!15}\underline{#1}}

\usepackage{hyperref}
\newcommand\blfootnote[1]{%
  \begingroup
    \let\thefootnote\relax\footnotetext{#1}%
  \endgroup
}

\title{FlowLong: Inference-time Long Video Generation via Manifold-constrained Tweedie Matching}

\author{
  Jangho Park$^{1,\ast}$ \quad
  Geon Yeong Park$^{1,\ast}$ \quad
  Gihyun Kwon$^{2,\dagger}$ \quad
  Jong Chul Ye$^{1,\dagger}$ \\
  $^{1}$KAIST \quad $^{2}$Amazon \\ \\
  \url{https://flowlong-video.github.io/}
}

\begin{document}

\maketitle
\blfootnote{$^{\ast}$Equal contribution. \quad $^{\dagger}$Co-corresponding authors.}

\begin{abstract}
Extending the generation horizon of video diffusion models to long sequences remains a long-standing and important challenge. Existing training-free approaches fall into two categories: extensions of bidirectional models, which are tightly coupled to specific architectures and suffer from quality degradation over long horizons, and autoregressive models, which accumulate drift errors due to exposure bias and tend to produce repetitive motion patterns. To address these issues, we propose a novel but simple inference-time approach for long video generation that is architecture-agnostic and requires no additional training. 
Our method generates long videos via overlapping sliding windows, where predicted clean samples from adjacent windows are blended via \emph{Tweedie matching} to enforce both \textbf{manifold constraint and temporal consistency} across overlap regions. \emph{Stochastic early-phase sampling} then synchronizes per-window trajectories by injecting fresh noise after each Tweedie matching correction in the high-noise phase, before transitioning to deterministic ODE sampling to preserve fine-grained visual fidelity.
Applied to various video generation models, our method generates videos several times longer than the native window length while outperforming both training-free and autoregressive baselines in temporal consistency and visual quality, and further extends to audio-video joint generation and text-to-3DGS without any fine-tuning.
\end{abstract}

\section{Introduction}
Video Diffusion Transformers (DiT)~\citep{dit} has driven remarkable progress in video generation, enabling models to produce videos of unprecedented fidelity and motion quality. This rapid advancement has further extended its reach into a diverse range of generative tasks, including camera-controlled video generation~\citep{yu2025trajectorycrafter, bai2025recammaster, reangle, inversecrafter} and 3D/4D generation~\citep{sv3d, vist3a, cat4d, 4real, zero4d}. Among these growing demands, the need for longer video content is particularly pressing across a wide range of applications, from cinematic content creation and interactive storytelling to embodied world models~\citep{worldmodel, genie, lingbot, swm} and immersive AR/VR experiences, where short clips are insufficient. Despite these demands, generating videos significantly longer than the training length remains a fundamental challenge. Most video diffusion models are trained on short clips due to the scarcity of large-scale, high-quality long video data, and directly applying them beyond their training length leads to severe quality degradation.

\begin{figure}
    \centering
    \includegraphics[width=1\linewidth]{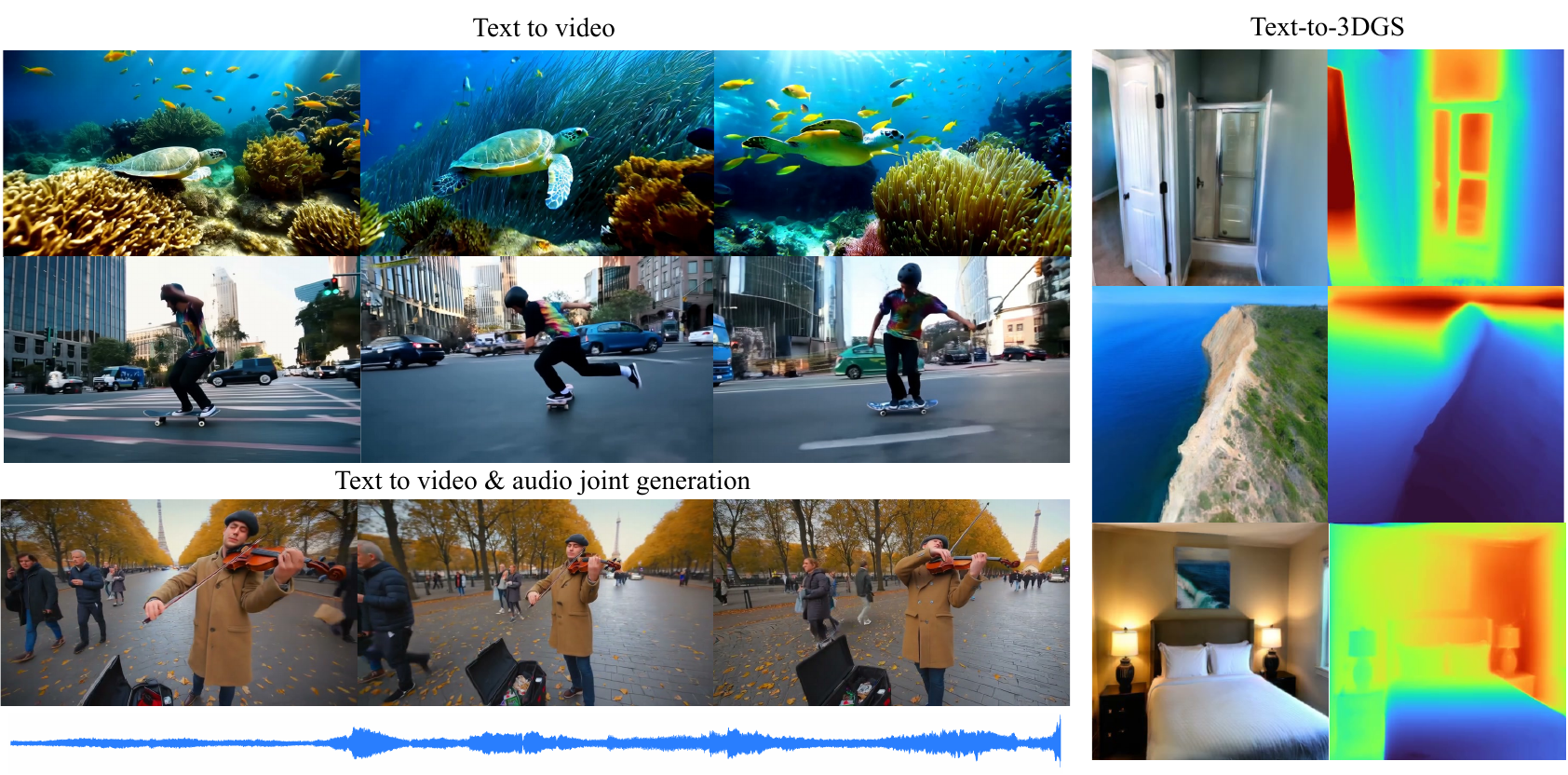}
    \vspace{-12pt}
    \caption{\textbf{Qualitative results.} \emph{FlowLong} extends pretrained video diffusion models beyond their native context window without any additional training. Being model-agnostic and training-free, it readily applies across diverse tasks: text-to-video generation, joint audio-video generation, and text-to-3DGS. This points toward a versatile, plug-and-play framework for extrapolating the generative horizon of any flow-based model. Please refer to the supplementary video.
    }
    \label{fig:placeholder}
    \vspace{-20pt}
\end{figure}

 This has motivated a growing body of work on long video generation, which falls into two categories. The first extends pre-trained bidirectional video diffusion models to longer sequences without additional training (e.g., FIFO-Diffusion~\citep{fifo}, RIFLEx~\citep{riflex}, UltraViCo~\citep{ultravico}). While these methods avoid additional training, they share limitations: consistency degrades as video length grows, visual artifacts accumulate over long horizons, and their reliance on architecture-specific modifications hinders applicability to new models. The second category formulates long video generation autoregressively. CausVid~\citep{causvid} demonstrates that distillation-based few-step generation can be applied to video, enabling autoregressive generation via KV-cache. Self-Forcing~\citep{selfforcing} further addresses the training-inference gap, inspiring follow-up works~\citep{selfforcingpp, rollingforcing, deepforcing, infinity-rope}. However, these methods suffer from several limitations. Reusing KV-cache across segments causes errors to accumulate over time, leading to \emph{exposure bias} and temporal drift. Motion diversity also degrades, as the model tends to produce repetitive motion patterns over long horizons. Furthermore, these approaches require distillation from a bidirectional teacher model, making them difficult to apply on the fly to recently introduced architectures such as joint audio-video models~\citep{ltx}.



To overcome these limitations, we propose a novel inference-time framework for long video generation, grounded in the geometric view of flow-based video generative models. Inspired by recent advances in diffusion-based inverse problem solvers~\citep{chung2023decomposed}, we reformulate long video generation as an inverse problem that aligns multiple chunk sampling trajectories toward a coherent sequence. Specifically, we regularize each chunk's denoising path with a guidance loss enforcing smooth and manifold-constrained transitions across overlap frames of adjacent chunks. This eventually reduces to a one-step gradient correction on the denoised estimate  during reverse sampling, which takes the closed-form of a simple per-frame interpolation on the overlap region—a procedure we call \emph{Tweedie matching}. To sustain the effect of this correction and prevent trajectories from reverting to their divergent ODE paths, we further propose \emph{stochastic early-phase sampling}: noise is injected during the initial stages to break ODE trajectory inertia and facilitate cross-chunk mixing, before transitioning to deterministic ODE sampling phase. Since our framework modulates only the sampling process, it is fully architecture-agnostic, training-free, and free from the exposure bias inherent in KV-cache reuse. It further extends seamlessly to audio-video joint generation and text-to-3D scene generation without any fine-tuning. Our contributions are as follows:

\begin{itemize}
    \item We propose \textbf{FlowLong}, a training-free, model-agnostic framework that extends pretrained flow-based diffusion models beyond their native generation horizon. Operating purely at inference time, FlowLong applies uniformly to text-to-video, audio-video joint, and text-to-3D scene generation without any architectural modification or fine-tuning.
    \item We propose \emph{Tweedie matching}, which enforces both manifold-constraint and temporal consistency by blending predicted clean samples across overlapping segments, and \emph{stochastic early-phase sampling}, which breaks per-window trajectory inertia by injecting stochastic noise in the high-noise regime before transitioning to deterministic ODE sampling.
    \item We validate FlowLong across text-to-video, audio-video joint generation, and text-to-3DGS, consistently outperforming both training-free and autoregressive baselines in qualitative and quantitative evaluations without any fine-tuning or backbone-specific modifications.
\end{itemize}

\section{Related Work}

\textbf{Bidirectional video diffusion.} 
Recent video diffusion models~\citep{wan, hunyuanvideo, ltx} adopt a bidirectional architecture that generates a fixed-length window of frames through full spatio-temporal attention. Training-free approaches extend these models to longer sequences via backbone-specific interventions: FIFO-Diffusion~\citep{fifo} denoises along a first-in-first-out queue with monotonically increasing noise levels, RIFLEx~\citep{riflex} reduces the intrinsic frequency of rotary positional embeddings to suppress temporal repetition, and UltraViCo~\citep{ultravico} concentrates attention by suppressing scores for tokens beyond the training window. All of these approaches depend on architecture-specific modifications, coupling them to particular backbones, and the quality still degrades as the target length grows beyond the training distribution. FlowLong instead leaves the backbone untouched and harmonizes multiple overlapping windows via Tweedie matching, decoupling video length from the native window size. 

\textbf{Autoregressive video diffusion.} The success of autoregressive approaches~\citep{causvid, selfforcing, selfforcingpp, rollingforcing, deepforcing} has demonstrated that fast and robust video generation is achievable through an autoregressive process, with pioneer works~\citep{causvid, selfforcing} extending distribution matching distillation (DMD)~\citep{dmd} to videos. Despite these advances, generating sequences beyond the trained length causes errors to accumulate, leading to drift and difficulty in maintaining global context coherence. Subsequent works address specific failure modes: Self-Forcing++~\citep{selfforcingpp} aligns training and inference through a rolling KV cache with backward noise initialization for over four-minute generation, Rolling Forcing~\citep{rollingforcing} mitigates exposure bias by training on the model's own histories with non-overlapping few-step distillation, FramePack~\citep{framepack} compresses past contexts by importance to bound the cache while planning sampling, and PFP~\citep{pfp} introduces a frame-query history encoder pretrained for dense temporal coverage and finetuned for content-level long-form consistency. Despite their differences, these methods share two structural limitations: every method depends on KV-cache reuse, leaving it susceptible to exposure bias, drift, and motion repetition over long horizons, and every method requires distillation from a bidirectional teacher, restricting applicability to architectures for which such a teacher already exists. In contrast, FlowLong samples all windows in parallel from independent Gaussian noise without KV-cache, eliminating exposure bias by construction and applying directly to architectures such as audio-video joint models~\citep{ltx} and text-to-3DGS models~\citep{vist3a}.

\section{Preliminaries}
\paragraph{Flow model.}
Flow matching~\citep{rectifedflow} defines a continuous normalizing flow that transports samples from a simple source distribution $p_1$ to a target distribution $p_0$ over $\mathbb{R}^d$ along a straight path.
For example, Rectified flow~\citep{rectifedflow} defines a linear interpolant between a data sample $\xb_0 \sim p_0$ and noise $\xb_1 \sim \Nc(\mathbf{0}, \mathbf{I})$:
\begin{equation}
    \xb_t = (1 - t)\xb_0 + t\xb_1.
\end{equation}

A neural network $\vb_\theta(\xb_t, t)$ is trained to approximate the velocity field $\vb(\xb_t) = \frac{d\xb_t}{dt}$ that transports $\mathbf{x}_1$ back to $\mathbf{x}_0$, via the following conditional flow matching objective:
\begin{equation}
\mathbb{E}_{t, \xb_0, \xb_1} \left[ \| \vb_\theta(\xb_t, t) - \vb(\xb_t|\xb_0) \|^2 \right],\quad  \vb(\xb_t|\xb_0)= \xb_1 - \xb_0
\end{equation}

\paragraph{Sampling.} Starting from $\mathbf{x}_1 \sim \mathcal{N}(\mathbf{0}, \mathbf{I})$, samples ($\xb_0 \sim p_0$) are generated by solving the learned ODE from $t=1$ to $t=0$:
\begin{equation}
\label{eq: ode}
    d\xb_t = v_\theta(\xb_t, t)\, dt.
\end{equation}

For example, an Euler step from time $t$ to $s < t$ reads
\begin{equation}
\label{eq: euler}
    \xb_s = \xb_t + (s-t) \vb_\theta(\xb_t, t).
\end{equation}
Defining the denoised and noisy estimates as
\begin{align}
\label{eq: denoised estimates}
    \hat{\xb}_{0|t} &:= \mathbb{E}[\xb_0 | \xb_t] = \xb_t - t \vb_\theta(\xb_t, t) \\
    \hat{\xb}_{1|t} &:= \mathbb{E}[\xb_1 | \xb_t] = \xb_t + (1-t) \vb_\theta(\xb_t, t) = \frac{\xb_t - (1-t) \hat{\xb}_{0|t}}{t},
\end{align}
which can also be equivalently derived from Tweedie's formula \citep{efron2011tweedie}. Then, an Euler step \eqref{eq: euler} can be reformulated as \citep{flowdps}:
\begin{equation}
\label{eq: euler_interp}
    \xb_s = (1-s) \hat{\xb}_{0|t} + s \hat{\xb}_{1|t},
\end{equation}
which corresponds to the interpolation between denoised and noisy estimates. For a text-guided flow model, the training objective is
often given by:
\begin{equation}
\mathbb{E}_{t, \xb_0, \xb_1} \left[ \| \vb_\theta(\xb_t, t, \cb) - \vb(\xb_t|\xb_0) \|^2 \right],
\end{equation}
where $\cb$ represents the textual embedding. Throughout this
paper, we will often omit $\cb$ from $\vb_\theta(\xb_t, t, \cb)$ or $\hat{\xb}_{0|t}(\cb) = \xb_t - t \vb_\theta(\xb_t, t,\cb)$ if it does not
lead to notational ambiguity. Following standard practice, we consider the latent flow model with a pretrained encoder-decoder $(\mathcal{E}_\phi, \mathcal{D}_\psi)$, and with a slight abuse of notation, continue to use $\xb$ to denote encoded video latents throughout.


\begin{figure}
    \centering
    \includegraphics[width=1\linewidth]{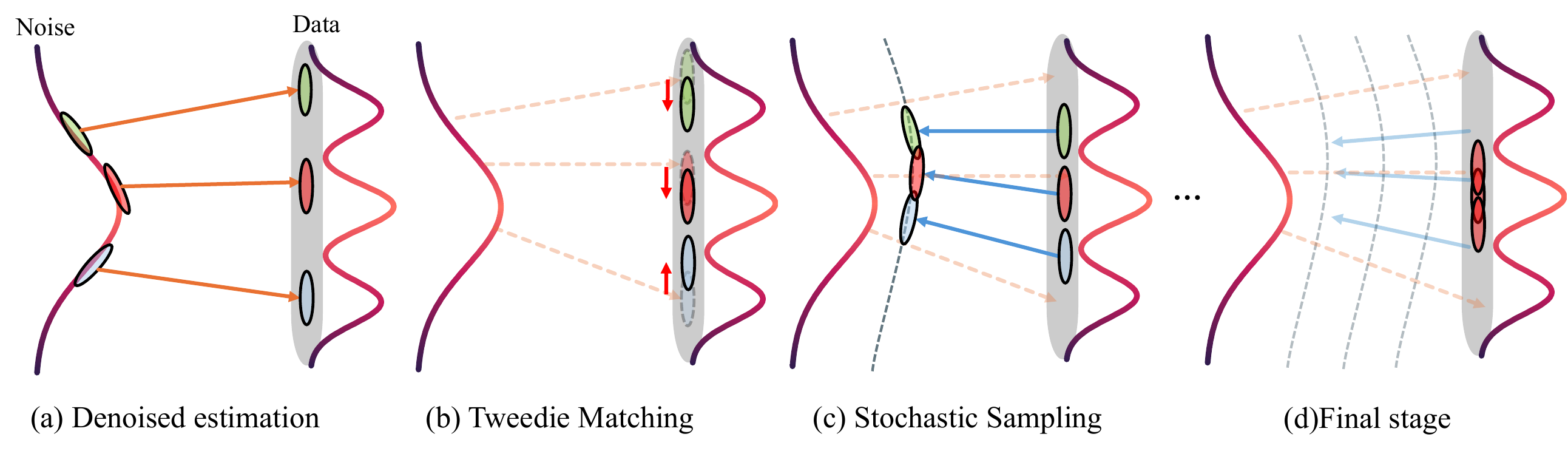}
    \vspace{-20pt}
    \caption{\textbf{Main pipeline.} We generate long videos in inference-time by harmonizing multiple short video chunk sampling trajectories. \textbf{(a)} Each chunk starts from independent noise and follows its own ODE, producing divergent denoised estimates. \textbf{(b)} We interpolate the denoised estimates on overlapping frames to align adjacent chunks (Sec \ref{sec: tweedie matching}). \textbf{(c)} In early sampling phase, we inject stochastic noise to prevent trajectories from reverting to their original divergent ODE paths (Sec \ref{sec: Stochastic Early-Phase Sampling}). \textbf{(d)} Repeating (b)--(c) progressively synchronizes all chunks into a coherent long video.
    }
    \label{fig:placeholder}
\end{figure}

\section{FlowLong: Inference-time Long Video Generation}
Pretrained video diffusion models~\citep{wan} learn the data distribution $p_0$ over $F$-frame video chunk latents $\xb \in \mathbb{R}^{F \times d}$ via flow matching. While these models produce high-quality short clips within the trained chunk length, they cannot natively extend videos longer than $F$ frames, restricting the scope of user interaction.

We address this limitation without fine-tuning. Given a pretrained model $\vb_\theta$, our goal is to generate a coherent long video sequence $\Xb =(\xb^{1}, \dots, \xb^{K})$ comprising $N > F$ frames by simultaneously sampling $K$ overlapping chunks and harmonizing them into a temporally consistent sequence. Note that $\vb_\theta$ is invoked independently on each chunk at every sampling step—never on the full sequence. Each video chunk $\xb^k$ is conditioned on its own text prompt $\cb_k$, which may vary across chunks. 

Towards training-free long video generation, the central challenge is: \emph{How to synchronize frame transitions across different chunk sampling trajectories that may diverge due to independent ODE noise initializations or potentially distinct prompts}? To answer this question, we adopt a fundamentally different strategy by formulating long video generation as an optimization problem. Specifically, our framework is based on neighbor-chunk conditioned latent optimization objective, which, when minimized during the reverse sampling process, progressively aligns each adjacent video chunks for smooth transitions. To prevent early divergence and facilitate mixing across trajectories, we further cast the initial sampling phase as an SDE by injecting stochastic noise, transitioning to deterministic ODE sampling in later stages. More details follows.

\subsection{Tweedie Matching}
\label{sec: tweedie matching}

For a coherent long video sequence generation, we impose the following constraint -- adjacent video chunk latents $\xb^k$ and $\xb^{k+1}$ should share an consistent overlap of $O$ frames: within this overlap window, the last $O$ frames of chunk $k$ should coincide with the first $O$ frames of chunk $k+1$. To formalize this constraint, let $\mathbf{1}_{\Omega_k}, \mathbf{1}_{\Omega'_{k+1}} \in \{0, 1\}^F$ denote the indicator vectors: 
\begin{equation}
    \mathbf{1}_{\Omega_k} = (\underbrace{0,\dots,0}_{F-O},\underbrace{1,\dots,1}_{O})^\top, \qquad
    \mathbf{1}_{\Omega'_{k+1}} = (\underbrace{1,\dots,1}_{O},\underbrace{0,\dots,0}_{F-O})^\top,
\end{equation}

which gives corresponding frame-selection matrices $M_k, M'_{k+1} \in \mathbb{R}^{O \times F}$ as follows:
\begin{equation}
    M_k = \bigl[\, 0_{O \times (F-O)} \;\big|\; I_O \,\bigr], \qquad
    M'_{k+1} = \bigl[\, I_O \;\big|\; 0_{O \times (F-O)} \,\bigr].
\end{equation}
Both map a chunk $\xb$ into the shared overlap window $\mathbb{R}^{O \times d}$, where $M_k^\top M_k = \text{diag}(\mathbf{1}_{\Omega_k})$ and $M^{\prime \top}_{k+1} M'_{k+1} = \text{diag}(\mathbf{1}_{\Omega'_{k+1}})$. Then, the hard overlap constraint reads:
\begin{equation}
\label{eq: hard constraint}
    M_k\, \xb_0^{(k)} = M'_{k+1}\, \xb_0^{(k+1)}, \qquad k = 1, \dots, K{-}1.
\end{equation}

\paragraph{Guidance loss.} We relax \eqref{eq: hard constraint} into a sampling guidance loss $\ell_k$ defined on the clean manifold. At time $t$ and $k$-th chunk, $\ell_k$ is defined as:
\begin{equation}
\label{eq:l_long}
    \ell_k(\xb; t) = \frac{1}{2} \bigl\lVert M_k\, \xb - M'_{k+1}\, \hat{\xb}_{0|t}^{(k+1)}(\cb_{k+1}) \bigr\rVert^2,
\end{equation}
where $\hat{\xb}_{0|t}^{k+1}(\cb_{k+1})$ refers to the clean estimate of adjacent chunk $k+1$ as in \eqref{eq: denoised estimates}, $\xb \in \Mc$ with a clean data manifold $\Mc$. This guidance loss represents an ideal overlap condition that neighboring video chunk latents should satisfy. This formulation is structurally identical to the inverse problem template $\frac{1}{2} \bigl\lVert \yb - \Ac\xb \bigr\rVert^2$ in diffusion inverse solvers \citep{chung2023decomposed}, with forward operator $\Ac = M_k$ and measurement $\yb = M'_{k+1}\hat{\xb}_{0|t}^{(k+1)}(\cb_{k+1})$ given by the neighboring chunk. 

\paragraph{Latent optimization.} Following diffusion inverse solvers (DDS \citep{chung2023decomposed}), we can now integrate the optimization step of $\ell_{k}$ in terms of denoised estimates $\hat{\xb}_{0|t}^{k}(\cb_{k})$, resulting in a modulated Euler step ($t \rightarrow s$):
\begin{align}
\label{eq: euler_ours}
    \bar{\xb}_{0|t}^{k}(\cb_{k}) &:= \hat{\xb}_{0|t}^{k}(\cb_{k}) - \gamma_t \nabla_{\hat{\xb}_{0|t}^{k}(\cb_{k})} \ell_{k}(\hat{\xb}_{0|t}^{k}(\cb_{k}); t) \nonumber \\
    \xb_s^k &= (1-s) \bar{\xb}_{0|t}^{k}(\cb_{k}) + s \bar{\xb}_{1|t}^{k}(\cb_{k}),
\end{align}
where $\bar{\xb}_{1|t}^{k}(\cb_{k}) = \frac{\xb_t - (1-t) \bar{\xb}_{0|t}(\cb_k)}{t}$ as in \eqref{eq: denoised estimates}. The gradient guidance is delineated as follows:

\begin{equation}
    \nabla_{\hat{\xb}_{0|t}^{(k)}} \ell_k(\hat{\xb}_{0|t}^{k}(\cb_{k}); t) = M_k^\top \bigl( M_k\, \hat{\xb}_{0|t}^{(k)}(\cb_k) - M'_{k+1}\, \hat{\xb}_{0|t}^{(k+1)}(\cb_{k+1}) \bigr),
\end{equation}
which is supported only on the overlap frames. Specifically, \eqref{eq: euler_ours} is reformulated as:

\begin{equation}
\label{eq: tweedie update}
    \bar{\xb}_{0|t}^{(k)} = \hat{\xb}_{0|t}^{(k)}(\cb_k) - \lambda\, M_k^\top \bigl( M_k\, \hat{\xb}_{0|t}^{(k)}(\cb_k) - M'_{k+1}\, \hat{\xb}_{0|t}^{(k+1)}(\cb_{k+1}) \bigr),
\end{equation}
where $\lambda > 0$ absorbs the step size $\gamma_t$. Per frame, since $M_k^\top M_k = \text{diag}(\mathbf{1}_{\Omega_k})$, this update reads

\begin{equation}
    \bar{\xb}_{0|t}^{(k)}[j] =
    \begin{cases}
        \hat{\xb}_{0|t}^{(k)}(\cb_k)[j], & j \notin \Omega_k, \\[4pt]
        (1{-}\lambda_j)\, \hat{\xb}_{0|t}^{(k)}(\cb_k)[j] + \lambda_j\, \hat{\xb}_{0|t}^{(k+1)}(\cb_{k+1})[j'], & j \in \Omega_k,
    \end{cases}
\end{equation}
where $j' = j - (F-O)$ is the corresponding frame index in $\Omega'_{k+1}$, and $\lambda_j$ refers to per-frame step size. Non-overlap frames ($j \notin \Omega_k$) remain untouched, while overlap frames are interpolated toward each neighbor's denoised estimate from Tweedie's formula. Thus, we call this update as  \emph{Tweedie matching}, which is manifold-constrained due to the use of DDS. A symmetric update is applied to chunk $k+1$ and others. In practice, we set $\lambda_j$ to a symmetric schedule over the overlap window, ensuring smooth frame-level blending and exact consistency at the boundary, so that each overlap region is stored once and shared by both chunks without duplication. Please refer to appendix for more details.

\begin{figure}
    \centering
    \includegraphics[width=1\linewidth]{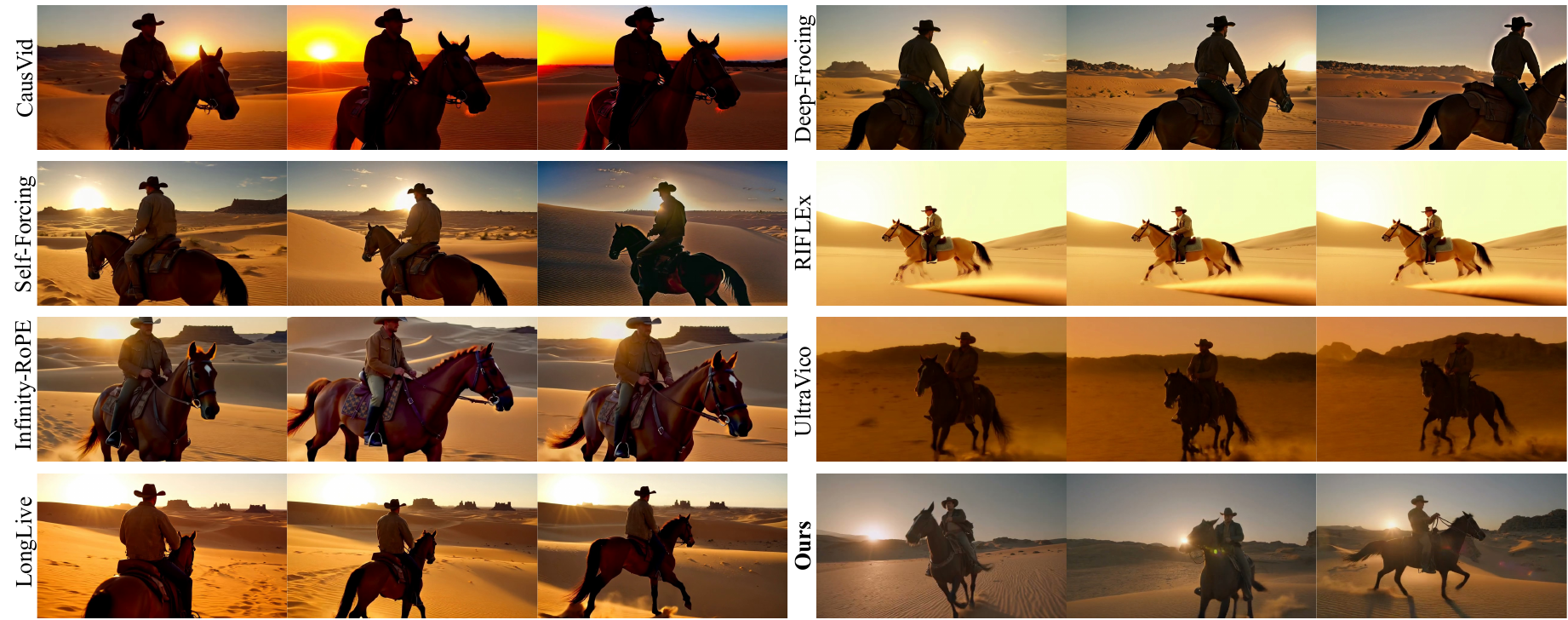}
    \caption{\textbf{Qualitative comparison of 30-second video generation.} Baselines suffer from repetitive motion patterns and drift errors caused by accumulated exposure bias. In contrast, our method produces videos with diverse motion dynamics and is robust to exposure bias. Please refer to the supplementary video.}
    \label{fig:qualitative_longvid}
\end{figure}

\paragraph{Prompt conditioning.} When all chunks share a common prompt ($\cb_k=\cb$ for all $k$), the guidance loss~\eqref{eq:l_long} enforces temporal coherence under a single scene description. For multi-shot generation with per-chunk prompts $\cb_1, \dots, \cb_K$, we condition each chunk on a shared global prompt $\cb_{\mathrm{global}}$ to maintain stylistic and semantic consistency across scene transitions, while the additional per-chunk prompt supplements local content.

\subsection{Stochastic Early-Phase Sampling}
\label{sec: Stochastic Early-Phase Sampling}
While Sec. \ref{sec: tweedie matching} regularizes the denoising paths toward a coherent long video sequence, under a deterministic ODE sampling regime, this correction may be insufficient to fully synchronize video chunks. Specifically, even after the clean estimate is pulled toward the neighbor via Tweedie matching, the deterministic renoising step drives $\xb_s^k$ back toward the original ODE trajectory. When ODE trajectories are initialized from independent Gaussian noise (and conditioned on potentially distinct prompts) their trajectories may be far apart in latent space, and this inertia prevents the long video harmonization across time steps.

To break this inertia, we inject stochastic noise during the early sampling phase by casting the renoising step in stochastic form. The injected noise perturbs each chunk away from its deterministic trajectory, effectively renoising the state after each Tweedie matching correction. Following FlowDPS~\citep{flowdps}, we mix the stochastic noise in \eqref{eq: euler_ours} as:

\begin{equation}
\label{eq: stochastic renoising}
    \xb_s^k = (1-s) \bar{\xb}_{0|t}^{k}(\cb_{k}) + s \tilde{\xb}_{1|t}^{k}(\cb_{k}),
\end{equation}
where 

\begin{equation}
\label{eq: noise injection}
 \tilde{\xb}_{1|t}^{k}(\cb_{k}) = \sqrt{1 - \eta_t} \bar{\xb}_{1|t}^{k}(\cb_{k}) + \sqrt{\eta_t} \epsilonb, \quad \epsilonb \sim \Nc(\mathbf{0}, \mathbf{I}).
\end{equation}

By setting $\kappa_{s,t} = s\sqrt{\eta_t}$, the renoising step in \eqref{eq: stochastic renoising} can be reformulated in stochastic form as follows: 
\begin{equation}
    \xb_s^k = (1-s) \bar{\xb}_{0|t}^{k}(\cb_{k}) + \sqrt{s^2 - \kappa_{s,t}^2} \bar{\xb}_{1|t}^{k}(\cb_{k}) + \sqrt{\kappa_{s,t}^2} \epsilonb,    
\end{equation}
which decomposes the renoising into a deterministic component along $\bar{\xb}_{1|t}^{k}(\cb_{k})$ and a stochastic perturbation of magnitude $\kappa_{s,t}$.

In practice, we adopt a binary schedule $\eta_t = \mathbbm{1}(t \geq t^*)$ for a threhold $t^*$. This implies that the early stochastic phase ($t \geq t^*$) uses full stochastic renoising to remix trajectories after each Tweedie matching correction, while the later phase $(t < t^*)$ reverts to deterministic ODE sampling to preserve fine-grained visual fidelity. As shown in Figure~\ref{fig:qualitative_longvid}, experimental results demonstrate that this hybrid sampling approach significantly improves temporal consistency and mitigates exposure bias in long video generation. Exploring smoother schedules for $\eta_t$ is an interesting direction for future work. 

\subsection{Extend to other generation tasks}
Our framework is not specific to temporal extension of visual video models; it applies broadly to any setting where a pretrained flow model generates fixed-size windows and the goal is to produce outputs that exceed this native horizon. The key requirement is that adjacent windows share an overlap region where Tweedie matching can enforce consistency. As promising examples, we demonstrate two additional applications: audio-video joint generation and text-to-3D generation. Crucially, \textit{none} of these extensions require \textit{fine-tuning}, in contrast to existing autoregressive long video models that must be retrained for each backbone and task.

\paragraph{{Audio-video joint generation.}} 
LTX-2~\citep{ltx} is a flow-matching video DiT augmented with an audio branch and cross-modal attention, denoising video and audio latents jointly under a shared text condition. To extend it beyond its native window, we decompose each modality into $M$ overlapping chunks aligned through the model's frame-rate ratio, and apply Tweedie matching (Sec.~\ref{sec: tweedie matching}) to both streams with the same overlap schedule $\lambda_j$. The corrected estimates are then advanced by stochastic early-phase renoising (Sec.~\ref{sec: Stochastic Early-Phase Sampling}) with independent perturbations $\epsilon^v, \epsilon^a$ per modality, producing arbitrarily long, phase-locked audio-video sequences without any fine-tuning.

\paragraph{{Text-to-3D generation.}} 
  VIST3A~\citep{vist3a} stitches a feed-forward 3D reconstructor, AnySplat~\citep{anysplat}, into the latent space of Wan~2.1~\citep{wan} via a lightweight bridge layer, converting a denoised video latent into 3D Gaussian splats in a single forward pass without per-scene optimization. To extend it beyond the native window, we initialize a noisy latent of the desired extrapolated length, decompose it into $M$ overlapping chunks, and apply Tweedie matching (Sec.~\ref{sec: tweedie matching}) followed by stochastic early-phase renoising (Sec.~\ref{sec: Stochastic Early-Phase Sampling}) at every sampling step. The resulting extended video latent is then decoded and fed to AnySplat, producing a longer 3D scene from text alone.



\begin{figure}
    \centering
    \includegraphics[width=1\linewidth]{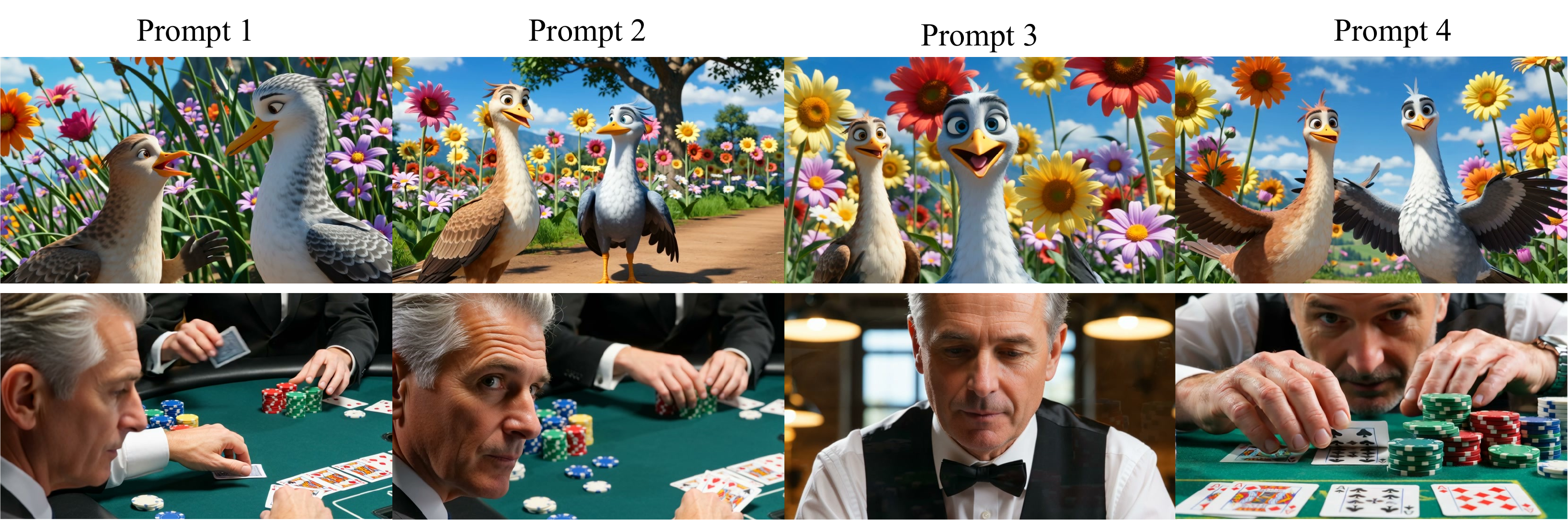}
    \caption{\textbf{Qualitative results of multi-prompt long video.} Our approach supports both global coherence via a shared prompt and fine-grained diversity via per-chunk local prompts.}
    \label{fig:multi_prompt}
\end{figure}

\section{Experiments}
For long video generation, we compare against bidirectional diffusion models (RIFLEx~\citep{riflex}, UltraViCo~\citep{ultravico}) and autoregressive diffusion models (CausVid~\citep{causvid}, Self-Forcing~\citep{selfforcing}, Deep-Forcing~\citep{deepforcing}, $\infty$
-RoPE~\citep{infinity-rope}, LongLive~\citep{longlive}), and against VIST3A~\citep{vist3a} for text-to-3DGS generation. We evaluate using VBench~\citep{vbench} across seven dimensions: aesthetic quality, imaging quality, background consistency, subject consistency, motion smoothness, dynamic degree, and temporal flickering, generating 30s and 60s videos from 100 MovieGen Bench~\citep{moviegen} prompts and 100 SceneBench~\citep{prometheus} prompts for 3DGS. Our method is applied without additional training on Wan~2.1-T2V-1.3B~\citep{wan} and LTX-2~\citep{ltx} for long video generation, and Wan~2.1-T2V-14B with AnySplat~\citep{anysplat} for text-to-3DGS, all on a single NVIDIA H100 GPU.

\begin{figure}
    \centering
    \includegraphics[width=1\linewidth]{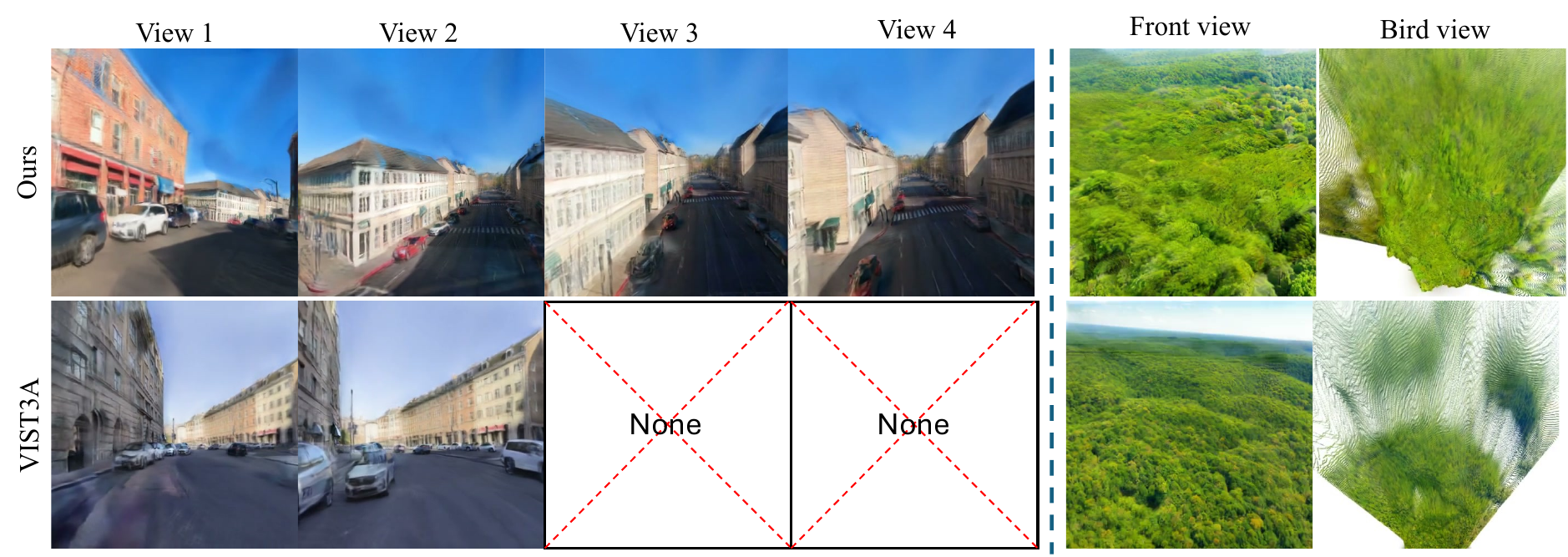}
    \caption{\textbf{Qualitative comparison of text to 3DGS generation.} These results demonstrate the rendered 3DGS generated from text. VIST3A~\citep{vist3a} is limited to the native context window of its pretrained video model, failing to produce views beyond this range. Our method extrapolates beyond this fixed capacity, generating 3DGS with substantially wider viewpoint coverage directly from text.} 
    \label{fig:qualitative_3dgs}
    \vspace{-10pt}
\end{figure}

\begin{table*}[t]
  \centering
  \caption{\textbf{Performance comparisons on 30s and 60s videos.} The \colorbox{blue!30}{first} and \colorbox{blue!15}{second} values
  are highlighted. We report VBench scores for each baseline, categorized by model type.}
  \label{tab:performance_comparison}
  \setlength{\tabcolsep}{2pt}
  \resizebox{\textwidth}{!}{%
  \begin{tabular}{l|cccccccc|cccccccc}
  \toprule
  \multirow{2}{*}{Model} & \multicolumn{8}{c|}{Results on 30s $\uparrow$} & \multicolumn{8}{c}{Results on 60s $\uparrow$} \\
  \cmidrule(lr){2-9} \cmidrule(lr){10-17}
  & \shortstack{Aesthetic \\ Quality} & \shortstack{Background \\ Consistency} & \shortstack{Dynamic \\ Degree} & \shortstack{Imaging
  \\ Quality} & \shortstack{Motion \\ Smoothness} & \shortstack{Subject \\ Consistency} & \shortstack{Temporal \\ Flickering} &
  Overall$\uparrow$
  & \shortstack{Aesthetic \\ Quality} & \shortstack{Background \\ Consistency} & \shortstack{Dynamic \\ Degree} & \shortstack{Imaging
  \\ Quality} & \shortstack{Motion \\ Smoothness} & \shortstack{Subject \\ Consistency} & \shortstack{Temporal \\ Flickering} &
  Overall$\uparrow$ \\
  \midrule
  \multicolumn{1}{l}{\textit{Bidirectional models -}1.3B} \\
  RIFLEx        & 0.44 & \best{0.97} & 0.08 & 0.41 & \best{0.99} & \best{0.97} & \best{0.99} & 0.6943 & - & - & - & - & - & - & - & -
  \\
  UltraViCo     & \second{0.4708} & \second{0.9348} & \second{0.5612} & \second{0.4334} & \second{0.9895} & \second{0.8793} &
  \second{0.9866} & \second{0.7508} & - & - & - & - & - & - & - & - \\
  Wan2.1 + Ours & \best{0.5777} & 0.9305 & \best{0.7800} & \best{0.6368} & 0.9877 & 0.8751 & 0.9753 & \best{0.8233} & - & - & - & - & -
   & - & - & - \\
  \midrule
  \multicolumn{1}{l}{\textit{Bidirectional models} - 14B} \\
  Ltx2          & \best{0.5412} & 0.8845 & \best{0.6251} & 0.6124 & 0.9813 & 0.8152 & 0.9482 & 0.7733 & - & - & - & - & - & - & - & -
  \\
  Ltx2 + Ours   & 0.5337 & \best{0.9016} & 0.6162 & \best{0.6393} & \best{0.9852} & \best{0.8201} & \best{0.9773} & \best{0.7812} & - &
   - & - & - & - & - & - & - \\
  \midrule
  \multicolumn{1}{l}{\textit{Autoregressive models}-1.3B} \\
  CausVid       & 0.5773 & 0.9037 & 0.4545 & 0.6556 & 0.9819 & 0.8874 & 0.9718 & 0.7760 & \second{0.5746} & 0.8871 & 0.4242 & 0.6438 &
  0.9815 & 0.8613 & 0.9723 & 0.7636 \\
  Self-Forcing  & 0.5523 & 0.9064 & 0.5455 & \second{0.6893} & 0.9858 & 0.8760 & 0.9752 & 0.7901 & 0.5355 & 0.8690 & 0.4747 & 0.6632 &
  0.9854 & 0.8056 & \second{0.9776} & 0.7587 \\
  Deep-Forcing  & 0.5667 & 0.9280 & \second{0.6566} & 0.6872 & 0.9836 & 0.9019 & 0.9718 & \second{0.8137} & 0.5691 & \second{0.9310} &
  \second{0.5253} & \second{0.6831} & 0.9850 & \second{0.9099} & 0.9745 & \second{0.7968} \\
  $\infty$-RoPE & 0.5724 & \second{0.9352} & 0.5102 & 0.6763 & 0.9870 & \second{0.9128} & \second{0.9765} & 0.7958 & 0.5640 & 0.9294 &
  0.5102 & 0.6803 & 0.9868 & 0.9046 & 0.9764 & 0.7931 \\
  LongLive      & \best{0.5868} & \best{0.9453} & 0.3535 & \best{0.6967} & \best{0.9895} & \best{0.9294} & \best{0.9792} & 0.7829 &
  \best{0.5885} & \best{0.9413} & 0.4141 & \best{0.6916} & \best{0.9891} & \best{0.9279} & \best{0.9784} & 0.7902 \\
  Wan2.1 + Ours & \second{0.5777} & 0.9305 & \best{0.7800} & 0.6368 & \second{0.9877} & 0.8751 & 0.9753 & \best{0.8233} & 0.5738 &
  0.9213 & \best{0.8200} & 0.6391 & \second{0.9869} & 0.8605 & 0.9738 & \best{0.8251} \\
  \bottomrule
  \end{tabular}%
  }
  \vspace{-10pt}
\end{table*}

\subsection{Long video generation}
\textbf{Qualitative results.} We provide a qualitative comparison of 30s video generation in Figure~\ref{fig:qualitative_longvid}. For bidirectional models~\citep{ultravico, riflex}, as the target video length increases beyond 30 seconds, meaningful motion nearly vanishes and pixel values become saturated. A similar phenomenon is observed in autoregressive models~\citep{causvid, selfforcing, infinity-rope, longlive, deepforcing}, where pixel values progressively saturate over time, leading to error drift. Furthermore, since these models continuously cache the key-value pairs of previous frames, the diversity of motion is severely limited, resulting in repetitive motion patterns. In contrast, our method regularizes and samples videos from independent initial points, which enables rich motion diversity and effectively eliminates the error drift that accumulates over time.

\textbf{Quantitative results.}
Table~\ref{tab:performance_comparison} reports VBench scores for 30s and 60s video generation, organized into three groups by model type. Against training-free bidirectional models (RIFLEx, UltraViCo), our model achieves the best overall score, with superior results including Dynamic Degree. For LTX2, where no training-free method supports generation beyond 30 seconds, we compare against a sliding-window baseline and achieve higher overall scores across most metrics. In the comparison with autoregressive models, our model outperforms all baselines by a substantial margin on Dynamic Degree, yielding the best overall performance and demonstrating robust motion diversity in long video generation. Additionally, as shown in Figure~\ref{fig:multi_prompt}, our method supports multi-shot generation by combining a global prompt with per-chunk local prompts, enabling diverse and semantically coherent scene transitions across extended video sequences.

\subsection{Text-to-3DGS}
\textbf{Qualitative results.} Fig.~\ref{fig:qualitative_3dgs} compares the baseline VIST3A with our extended pipeline on scene-level 3D generation. Since the baseline is restricted to the fixed generation window of the pre-trained video model, both the number and the spatial coverage of the resulting 3D Gaussian splats remain limited. Our method, by contrast, extrapolates the video latent beyond the native window through Tweedie matching and stochastic early-phase renoising, directly producing a longer video that translates into a substantially larger set of 3D Gaussians. As a result, our approach reconstructs noticeably longer 3D worlds with richer viewpoint diversity. The geometric benefit is also visible from a bird's-eye view: the baseline produces sparse Gaussians with limited spatial coverage, while ours yields a much denser point cloud that better represents the 3D scene geometry.

\begin{figure}
    \centering
    \includegraphics[width=1.0\linewidth]{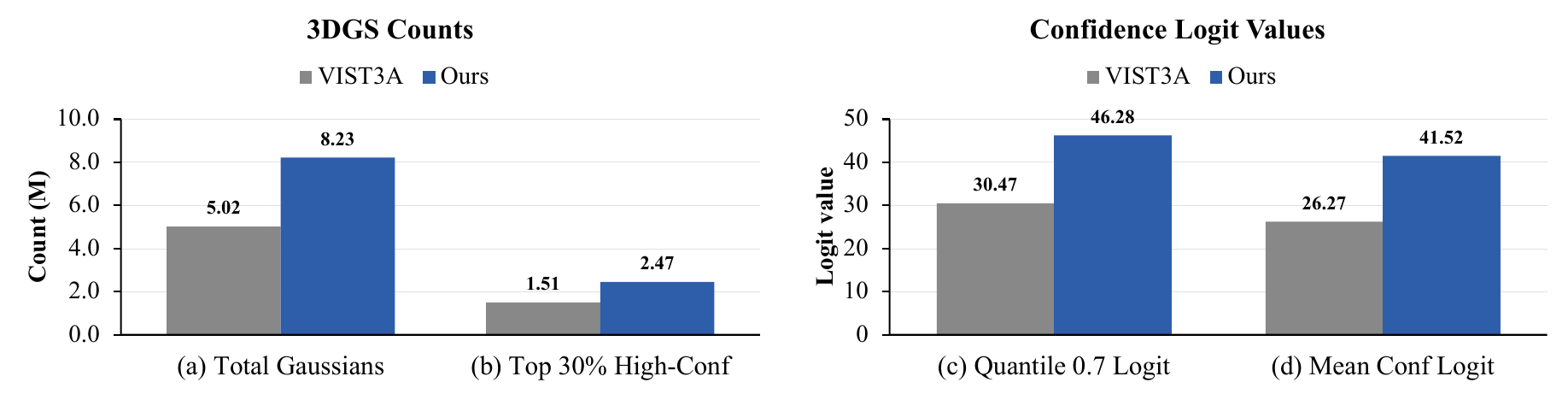}
    \vspace{-20pt}
    \caption{\textbf{Quantitative comparison of generated 3DGS.} (a)~Total number of Gaussians produced per scene. (b)~Number of Gaussians remaining after discarding the bottom 70\% by confidence. (c)~Average depth confidence logit across all Gaussians. (d)~Confidence cutoff above which the top-30\% most reliable Gaussians lie.}
    \label{fig:quantitative_3dgs}
    \vspace{-15pt}
\end{figure}

\textbf{Quantitative results.} AnySplat, the 3D Gaussian generator used in VIST3A, predicts a per-pixel depth confidence score indicating how reliable the estimated geometry is. We use this score to evaluate the quality of the generated 3D Gaussians and report all numbers averaged over 100 prompts from SceneBench. As shown in Fig.~\ref{fig:quantitative_3dgs}, our method generates 1.64$\times$ more Gaussians per scene than the baseline, thanks to the longer video sequence, and filtering to the top-30\% by confidence, our method still retains 2.47M Gaussians on average. The confidence scores further confirm this trend: the mean confidence logit rises from 26.27 (baseline) to 41.52 (ours), and the 0.7-quantile logit increases from 30.47 to 46.28, demonstrating that our method produces not only more but also higher-quality 3D Gaussians across diverse scenes.

\begin{figure}
    \centering
    \includegraphics[width=1\linewidth]{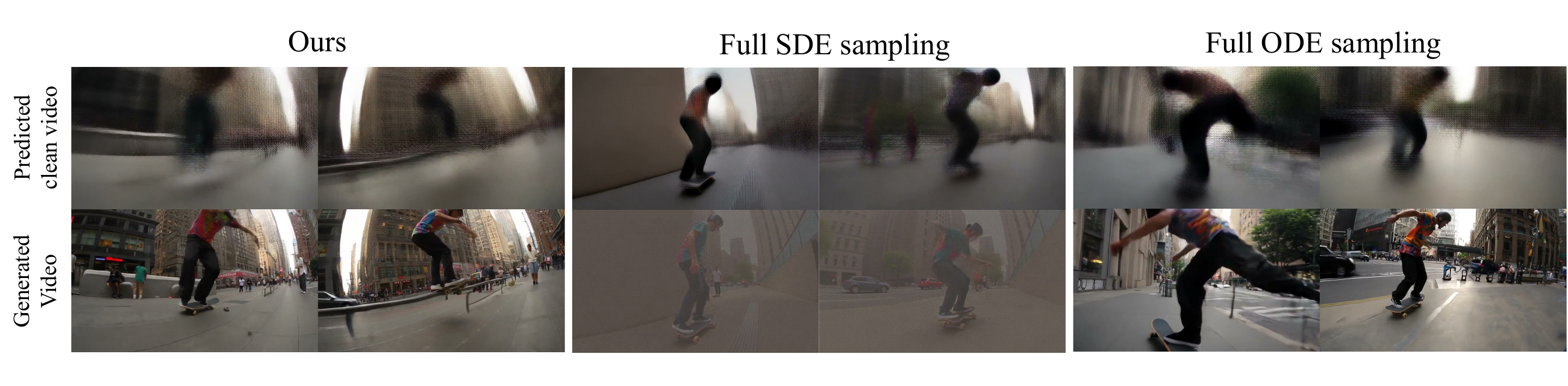}
    \vspace{-25pt}
    \caption{\textbf{Ablation result.} Comparison of denoised estimates and final generations across sampling strategies. The spatial layout is largely determined in the early sampling stages (Top row). Full SDE ($\eta_t=1$ in \eqref{eq: noise injection}) preserves temporal consistency across chunks but degrades visual quality, while full ODE ($\eta_t=0$) yields sharper results at the cost of exposure bias and temporal inconsistency.}
    \label{fig:ablation_sdeode}
    \vspace{-10pt}
\end{figure}

\subsection{Ablation}

\begin{wraptable}{r}{0.35\textwidth}
\centering
\vspace{-10pt}
\setlength{\tabcolsep}{4pt}
\resizebox{0.35\textwidth}{!}{%
\begin{tabular}{l|ccc}
\toprule
Method & Consistency $\uparrow$ & Motion $\uparrow$ & Quality $\uparrow$ \\
\midrule
Full SDE sampling & 0.9427 & 0.9449 & 0.5298 \\
Full ODE sampling & \second{0.9604} & 0.9621 & \second{0.6075} \\
$x_t$ matching   & 0.9579 & \best{0.9690} & 0.5862 \\
\midrule
Ours             & \best{0.9615} & \second{0.9685} & \best{0.6359} \\
\bottomrule
\end{tabular}%
}
\caption{\textbf{Ablation study.} The \colorbox{blue!30}{first} and \colorbox{blue!15}{second} values are highlighted. Our method achieves the best overall performance.}
\label{tab:ablation}
\vspace{-6pt}
\end{wraptable}

\vspace{-6pt}
We conduct ablation studies on two key components of our method. First, we ablate the Tweedie matching strategy for handling overlapping regions between segments. As shown in Table~\ref{tab:ablation}, compared to blending overlapping regions at an arbitrary noise level $t$~\citep{multidiffusion}, our method of matching overlaps in the predicted clean sample space achieves higher scores across Dynamic Degree, Consistency, and Quality. Second, we ablate the Stochastic Early-Phase Sampling strategy by comparing against full-SDE and full-ODE sampling. Our hybrid approach outperforms both alternatives, combining the high image quality of ODE and the temporal consistency of SDE. As shown in Figure~\ref{fig:ablation_sdeode}, ODE sampling produces frames that appear independent, while our hybrid approach and full-SDE maintain temporal consistency, with our method additionally preserving the quality of ODE.

\vspace{-6pt}
\section{Conclusion}
We presented FlowLong, a training-free, architecture-agnostic framework for long video generation built on two core components: Tweedie matching, which enforces temporal consistency across overlapping windows by blending predicted clean samples, and stochastic early-phase sampling, which synchronizes per-window trajectories by injecting stochastic noise in the high-noise regime before transitioning to deterministic ODE sampling. These components address the key failure modes of prior approaches without any architectural modifications or additional training. We validated FlowLong across text-to-video, audio-video joint generation, and text-to-3DGS tasks, consistently outperforming both training-free and autoregressive baselines. One limitation is that our overlap-based consistency constraint is inherently local, which may hinder global semantic coherence in extremely long videos, and we leave this as future work.

\bibliographystyle{plain}
\newpage

\bibliography{main}
\newpage
\appendix

\section{Tweedie Matching: Implementation Details}
\label{app: tweedie matching}

This appendix expands the practical details of Sec.~\ref{sec: tweedie matching}.
We describe the latent-space window geometry (\S\ref{app: geometry}), the explicit form of the per-frame blending schedule $\lambda_j$ (\S\ref{app: lambda}), an equivalence between iterating Eq.~\eqref{eq: tweedie update} over all chunk pairs and a single weighted aggregation (\S\ref{app: reduction}), the aggregation algorithm used in our implementation (\S\ref{app: aggregation}), and the corresponding geometry for audio--video joint generation (\S\ref{app: audio geometry}).

\subsection{Window geometry in latent space}
\label{app: geometry}

All operations are performed in the latent space of a video VAE with temporal stride $r$ ($r=8$ for LTX-2). We denote
\begin{itemize}
    \item $F$: number of latent frames per chunk (window length);
    \item $O$: number of latent frames in the \emph{blending zone}---the last $O$ frames of each chunk, on which Tweedie matching is applied;
    \item $S$: stride between the global start indices of consecutive chunks;
    \item $K$: number of chunks (consistent with Sec.~\ref{sec: tweedie matching});
    \item $N = F + (K-1)\,S$: total number of latent frames in the long-video buffer.
\end{itemize}
We require
\begin{equation}
\label{eq: geometry constraint}
    O \;\geq\; S,
\end{equation}
so that every latent frame in chunk $k$'s blending zone is also predicted by chunk $k+1$ at the same global temporal position; otherwise $M^{\prime}_{k+1}\hat{\xb}_{0|t}^{(k+1)}$ in the guidance loss \eqref{eq:l_long} would refer to frames that chunk $k+1$ never observed.

In practice the user specifies the pixel-space window size $W$ and the pixel index $w$ at which the overlap region begins inside a window. The latent quantities are then obtained by
\begin{equation}
\label{eq: pixel to latent}
    F = \tfrac{W-1}{r} + 1,
    \qquad S = \bigl\lfloor (W - w)/r \bigr\rfloor,
    \qquad O = F - \lfloor w/r \rfloor.
\end{equation}
For the LTX-2 backbone with $W=121$ and $w=64$, this yields $(F, O, S)=(16, 8, 7)$, satisfying \eqref{eq: geometry constraint} with one frame of slack; we use this configuration throughout our text-to-video experiments.

We index the global latent buffer as $g \in \{0,\dots,N-1\}$ and the local frames within chunk $k\in\{1,\dots,K\}$ as $j \in \{0,\dots,F-1\}$, related by $g = (k-1)\,S + j$. Chunk $k$'s blending zone in local indexing is therefore
\begin{equation}
    \Omega_k \;=\; \{\,F-O,\; F-O+1,\; \dots,\; F-1\,\},
\end{equation}
in agreement with the indicator vector $\mathbf{1}_{\Omega_k}$ in Sec.~\ref{sec: tweedie matching}.

\subsection{The per-frame blending schedule \texorpdfstring{$\lambda_j$}{lambda j}}
\label{app: lambda}

On the blending zone, Eq.~\eqref{eq: tweedie update} reduces to a per-frame convex combination between the two adjacent clean estimates with weight $\lambda_j$. We adopt the linear schedule
\begin{equation}
\label{eq: lambda schedule}
    \lambda_j \;=\; \frac{j-(F-O)}{O-1},    \qquad j \in \Omega_k,
\end{equation}
which has three properties used throughout the rest of the appendix.
\begin{enumerate}
    \item \textbf{Boundary consistency.} $\lambda_{F-O}=0$ and $\lambda_{F-1}=1$.    Hence the leftmost frame of the blending zone is taken entirely from chunk $k$, and the rightmost frame entirely from chunk $k+1$. The matched estimate $\bar{\xb}_{0|t}^{(k)}$ therefore agrees exactly with $\hat{\xb}_{0|t}^{(k)}$ at the seam $j=F-O$ and exactly with $\hat{\xb}_{0|t}^{(k+1)}$ at $j=F-1$, eliminating any discontinuity at either side of the overlap.
    \item \textbf{Symmetry.} Setting the local index $i = j-(F-O)\in\{0,\dots,O-1\}$, we have $\lambda_j = i/(O-1)$ and the mirror identity
    \begin{equation*}
        \lambda_j + \lambda_{j_{\mathrm{mir}}} \;=\; 1,        \qquad j_{\mathrm{mir}} \;=\; (F-O) + (O-1) - i \;=\; F-1-i.
    \end{equation*}
    Equivalently, chunk $k$'s update applies weight $\lambda_j$ to chunk $k+1$'s prediction, while the symmetric update applied to chunk $k+1$ (Sec.~\ref{sec: tweedie matching}) applies weight $1-\lambda_j$ to chunk $k$'s prediction at the same global frame. Both updates therefore produce the \emph{same} convex combination $(1-\lambda_j)\,\hat{\xb}_{0|t}^{(k)}[j] + \lambda_j\,\hat{\xb}_{0|t}^{(k+1)}[j']$. We exploit this in \S\ref{app: reduction} to compute the aggregation only once.
    \item \textbf{Smoothness.} $\lambda_j$ is linear in the frame index, so frame-level transitions across the blending zone are uniform. We did not observe additional gains from smoother schedules (e.g.\ raised-cosine windows) in preliminary experiments, while the linear form admits the simple equivalence in \S\ref{app: reduction}.
\end{enumerate}

\subsection{Multi-chunk pairwise updates collapse to a single weighted aggregation}
\label{app: reduction}

Sec.~\ref{sec: tweedie matching} states the modulated Euler step \eqref{eq: euler_ours} for an adjacent pair $(k,k+1)$. With $K$ chunks one would, in principle, iterate this update over all pairs $(1,2),(2,3),\dots,(K-1,K)$ and average where blending zones overlap. Under the symmetric linear schedule \eqref{eq: lambda schedule}, this iteration collapses into a single pass that produces, for each global frame $g$, a weighted average of \emph{exactly one} adjacent pair of clean predictions.

Partition the global indices $\{0,\dots,N-1\}$ as
\begin{itemize}
    \item the \emph{leading prefix} $g \in [0,\, F-O)$, owned solely by chunk $1$;
    \item per-pair \emph{blending zones} $\mathcal{B}_k \;=\; \bigl[\,(k-1)S + (F-O),\;\, (k-1)S + F\,\bigr)$ for $k=1,\dots,K-1$;
    \item the \emph{trailing suffix} $g \in [(K-1)S + (F-O),\, N)$, owned solely by chunk $K$;
    \item (only when $S>O$) an \emph{interior gap} $\mathcal{G}_k = [(k-1)S+F,\; kS+(F-O))$ between $\mathcal{B}_k$ and $\mathcal{B}_{k+1}$, owned solely by chunk $k+1$.
\end{itemize}
For $g \in \mathcal{B}_k$, both chunks $k$ and $k+1$ have a clean prediction at $g$, located at local indices $j = g-(k-1)S$ in chunk $k$ and $j' = j - S$ in chunk $k+1$; the linear schedule \eqref{eq: lambda schedule} assigns weight $1-\lambda_j$ to chunk $k$ and $\lambda_j$ to chunk $k+1$. By Property~2 of \S\ref{app: lambda}, the symmetric update for chunk $k+1$ produces the same convex combination, so the two pairwise updates can be replaced by a single write of the convex combination into the global buffer.

When $S = O$ the blending zones $\{\mathcal{B}_k\}_{k=1}^{K-1}$ tile the global buffer cleanly. When $S < O$ (the typical case in our experiments, e.g.\ $(F,O,S) = (16,8,7)$), $\mathcal{B}_k$ and $\mathcal{B}_{k+1}$ overlap by $O-S$ frames; we resolve the conflict by using the \emph{rightmost} pair, i.e.\ frames in $\mathcal{B}_k \cap \mathcal{B}_{k+1}$ are blended between chunks $k+1$ and $k+2$ (last-writer-wins; \S\ref{app: aggregation}). This preserves the boundary-consistency property of \S\ref{app: lambda}: at every seam, the long-video latent is identical to one of the two adjacent chunk predictions, so each overlap frame is stored exactly once and shared between adjacent chunks, as stated in Sec.~\ref{sec: tweedie matching}.

\subsection{Aggregation in the long-video buffer}
\label{app: aggregation}

The reduction in \S\ref{app: reduction} is realised by a single pass over the chunks. We allocate the long-video buffer $\hat{\Xb}_{0|t}\in\mathbb{R}^{N\times d}$ and write each global index exactly once: (i) the leading prefix $g\in[0,\,F-O)$ is copied from chunk $1$; (ii) for each $k=1,\dots,K-1$, the per-pair blending zone $\mathcal{B}_k$ is filled with $(1-\lambda_j)\hat{\xb}_{0|t}^{(k)}[j] + \lambda_j\hat{\xb}_{0|t}^{(k+1)}[j-S]$ for $j\in\Omega_k$, and any interior gap $\mathcal{G}_k$ that arises when $S>O$ is copied from chunk $k+1$; (iii) the trailing suffix $g\in[(K-1)S+(F-O),\,N)$ is copied from chunk $K$. Because the writes use direct assignment, no weight accumulation or re-normalisation is required, and overlaps between adjacent blending zones (the $S<O$ case in \S\ref{app: reduction}) are resolved by last-writer-wins. After the aggregation, the next-state update of Sec.~\ref{sec: Stochastic Early-Phase Sampling} (deterministic Euler in the low-noise phase, stochastic renoising in the high-noise phase) is applied directly on the global buffer $\hat{\Xb}_{0|t}$, and the resulting $\Xb_s$ is then re-sliced into $K$ overlapping windows for the next denoising step.

\subsection{Audio--video joint geometry}
\label{app: audio geometry}

LTX-2~\citep{ltx} jointly denoises a video latent and an audio latent under a shared text condition (Sec.~\ref{sec: tweedie matching} and Sec.~\ref{sec: Stochastic Early-Phase Sampling}). The two streams have different temporal rates: the video latent rate is $\mathrm{fps}_v / r$ frames per second, while the audio latent rate is $\rho_a$ latents per second ($\rho_a = 25$ for LTX-2). To keep Tweedie matching temporally consistent across both modalities we choose the audio geometry that matches the video stride in seconds.

Given the video pixel window size $W$, the pixel overlap-start index $w$, and the video frame rate $\mathrm{fps}_v$, the audio chunk length, stride, and overlap (all in audio latents) are
\begin{align}
    F_a \;&=\; \mathrm{round}\!\left(\frac{W}{\mathrm{fps}_v}\,\rho_a\right), \nonumber \\
    \label{eq: audio geometry}
    S_a \;&=\; \mathrm{round}\!\left(\frac{W - w}{\mathrm{fps}_v}\,\rho_a\right),
    \qquad
    O_a \;=\; F_a - S_a,
\end{align}
and the total audio length is $N_a = F_a + (K-1)\,S_a$. With $(W, w, \mathrm{fps}_v, \rho_a) = (121, 64, 24, 25)$ this gives $(F_a, O_a, S_a) = (126, 67, 59)$, which satisfies the same constraint $O_a \geq S_a$ as the video geometry \eqref{eq: geometry constraint}.

The continuous-time rounding in \eqref{eq: audio geometry} can introduce a misalignment of at most one audio latent ($\approx 1/\rho_a$ s, i.e.\ $40\,\mathrm{ms}$ for LTX-2) between the audio blending zone and the geometric overlap with the next chunk; we clamp this offset to zero in the aggregation, which we found to have no observable effect on audio--video phase locking. With this geometry, the aggregation of \S\ref{app: aggregation} is run independently on the video and audio buffers using their respective $(F,O,S)$, and stochastic early-phase sampling (Sec.~\ref{sec: Stochastic Early-Phase Sampling}) is then applied with separate noise samples $\epsilonb^v$ and $\epsilonb^a$, so per-modality trajectories synchronise across windows while remaining temporally aligned with each other through the shared chunk geometry.

\end{document}